\documentclass[11pt]{article}

\usepackage{amsmath,amsfonts,bm}

\def\eqref#1{equation~\ref{#1}}

\def\1{\bm{1}}

\DeclareMathAlphabet{\mathsfit}{\encodingdefault}{\sfdefault}{m}{sl}
\SetMathAlphabet{\mathsfit}{bold}{\encodingdefault}{\sfdefault}{bx}{n}

\usepackage[T1]{fontenc}

\usepackage[hyperindex,breaklinks]{hyperref}
\usepackage{multibib}
\usepackage{graphicx}
\usepackage{tabularx}
\usepackage{soul}
\usepackage{authblk}
\usepackage{epstopdf}
\usepackage[latin1]{inputenc}
\usepackage{booktabs}
\usepackage{amsmath}
\usepackage{longtable}
\usepackage{xstring}
\usepackage{microtype}
\usepackage{linguex}
\usepackage{enumitem}
\usepackage{todonotes}
\usepackage{xpatch}
\usepackage{float}
\usepackage{lrec2022}
\usepackage{times}
\usepackage{url}
\usepackage{latexsym}

\usepackage{authblk}

\hypersetup{
   breaklinks=true,   
}
\usepackage[hyphenbreaks]{breakurl}

\title{A Pragmatics-Centered Evaluation Framework\\ for Natural Language Understanding}

\name{Damien Sileo$^{3,2}$, Tim Van de Cruys$^{2}$, Camille Pradel$^{1}$, Philippe Muller$^{3}$}

\address{1:Synapse D\'eveloppement, 2:KU Leuven, 3:IRIT (University of Toulouse) \\
         damien.sileo@kuleuven.be\\}
\vspace{-2ex}

\abstract{
New models for natural language understanding have recently made an unparalleled amount of progress, which has led some researchers to suggest that the models induce {\it universal} text representations.
However, current benchmarks are predominantly targeting semantic phenomena;  we make the case that pragmatics needs to take center stage in the evaluation of natural language understanding.
We introduce PragmEval, a new benchmark for the evaluation of natural language understanding, that unites 11 pragmatics-focused evaluation datasets for English.
PragmEval can be used as supplementary training data in a multi-task learning setup, and is publicly available, alongside the code for gathering and preprocessing the datasets.
Using our evaluation suite, we show that natural language inference, a widely used pretraining task, does not result in genuinely universal representations, which presents a new challenge for multi-task learning.
}

\begin{document}
\maketitleabstract

\section{Introduction}
Over the last few years, pretrained for natural language understanding
(NLU) have made a remarkable amount of progress on a number of widely
accepted evaluation benchmarks. The GLUE benchmark~\cite{wang-etal-2018-glue}, for example, was designed to be a set of
challenging NLU tasks, such as question answering, sentiment analysis,
and textual entailment; yet, current state of the art systems surpass
human performance estimates on the average score of its subtasks~\cite{yang2019xlnet}. Similarly, the NLU subtasks that are part of
the SentEval~\cite{Conneau2017} framework, a widely used benchmark for the evaluation of
sentence-to-vector encoders, are successfully dealt with by current
neural models, with scores that exceed the 90\%
mark.\footnote{\href{http://nlpprogress.com/english/semantic_textual_similarity.html}{\texttt{http://nlpprogress.com/english/}}}

The results on these benchmarks are impressive, but sometimes lead to excessive optimism regarding the ability of current NLU models. For example, based on the resulting performance on the
above-mentioned benchmarks, a considerable number of researchers has even put
forward the claim that their models induce {\it universal}
representations~\cite{cer2018universal,kiros-chan-2018-inferlite,subramanian2018learning,wieting2016iclr,liu2019mt-dnn-kd}. It
is important to note, however, that benchmarks like SentEval and GLUE
are primarily focusing on semantic aspects, i.e. the literal and
uncontextualized content of text. While the semantics of language is
without doubt an important aspect of language, we believe that a
single focus on semantic aspects leads to an impoverished model of
language. For a versatile model of language, other aspects of language,
viz. pragmatic aspects, equally need to be taken into
account. Pragmatics focuses on the larger context that surrounds a
particular textual instance, and it is of vital importance for meaning
representations that aspire to lay a claim to universality. Consider
the following utterance :

\ex. You're standing on my foot.\label{ex:foot}

The utterance in~\ref{ex:foot} has a number of direct implications
that are logically entailed, such as the
implication that the hearer is standing on a body part of the speaker,
or the implication that the speaker is touching the hearer. But there
are also more indirect implications, that are not literally expressed,
but need to be inferred from the context, such as the implication that
the speaker wants the hearer to move away from them. The latter kind
of implication, that is indirectly implied by the context of an
utterance, is called {\it implicature}---a term coined by~\newcite{grice1975logic}. In real world applications, recognizing the
implicatures of a statement is arguably more important than
recognizing its mere semantic content.

The implicatures that are conveyed by an utterance are highly
dependent on its illocutionary force~\cite{austin1975things}. In
Austin's framework, the {\it locution} is the literal meaning of an
utterance, while the {\it illocution} is the goal that  the
utterance tries to achieve. When we restrict the meaning of
\ref{ex:foot} to its locution, the utterance is reduced to the mere
statement that the hearer is standing on the speaker's foot. However,
when we also take its illocution into account, it becomes clear
that the speaker actually formulates the request that the speaker step
away. The utterance's illocution is clearly an important part of the entire
meaning of the utterance, that is complementary to the literal content~\cite{GreenForceAndContent}.\footnote{In order to precisely determine
  their illocution, utterances have been categorized into classes
  called speech acts~\cite{searle1980speech}, such as {\sc
    Assertion}, {\sc Question} or {\sc Order} which have different
  kinds of effects on the world. For instance, constative speech acts
  (e.g. {\it the sky is blue}) describe a state of the world and are
  either true or false while performative speech acts (e.g. {\it I
    declare you husband and wife}) can change the world upon utterance~\cite{austin1975things}.}

The example above makes clear that pragmatics is a fundamental aspect
of the meaning of an utterance. Semantics focuses on the literal
content of utterances, but not on the kind of goal the speaker is
trying to achieve. Pragmatics and discourse tasks focus on the
actual use of language, so a pragmatics-centered evaluation could {\it
  by construction} be a better fit to evaluate how NLU models perform
in practical use cases---and in any case it should at least be used as a complement to
semantics-focused evaluation benchmarks. Ultimately, many use cases
of NLP models are related to conversations with end users or analysis
of structured documents. In such cases, discourse analysis (i.e. the
ability to parse high-level textual structures that take into account
the global context) is a prerequisite for human level
performance. Moreover, standard benchmarks often strongly influence
the evolution of NLU models, which means they should be as exhaustive
as possible, and closely related to the models' end use cases.


In this work, we compile a list of eleven pragmatics-focused tasks for English that are meant to complement existing benchmarks. We propose: (i) a new evaluation benchmark, named {\it PragmEval}, which is publicly available; \footnote{\url{https://github.com/synapse-developpement/PragmEval} and \url{https://huggingface.co/datasets/pragmeval}}
(ii) derivations of human accuracy estimates for some of the tasks; (iii) evaluation on these tasks of a state of the art generalizable NLU  model, viz. BERT (both with and without auxiliary finetunings); (iv) new comparisons of discourse-based and natural language inference based training signals, showing that the most widely used auxiliary finetuning dataset (MNLI) is not the best performing on PragmEval, which suggests a margin for improvement.

%





\section{Related Work}

Evaluation methods of NLU have been the object of heated debates since the proposal of the Turing Test. Automatic evaluations relying on annotated datasets are arguably limited but they have become customary practice. 
A popular method of evaluation is to predict sentence similarity~\cite{agirre2012semeval}, leveraging human annotated scores of similarity between sentence pairs. This task requires some representation of the sentences' semantic content beyond their surface form, and sentence similarity estimation tasks can potentially encompass many aspects. However, it is not clear how human annotators weigh semantic, stylistic, and discursive aspects while rating.

Using a set of  more focused and clearly defined tasks has been another popular approach. \newcite{Kiros2015} proposed a set of tasks and tools for sentence understanding evaluation. These thirteen tasks were compiled in the SentEval~\cite{Conneau2017} evaluation suite designed for automatic evaluation of pre-trained sentence embeddings. SentEval tasks are mostly based on sentiment analysis, semantic sentence similarity and natural language inference. Since SentEval evaluates sentence embeddings, the users have to provide a sentence encoder that is not fine-tuned during the evaluation.

 GLUE~\cite{wang-etal-2018-glue} proposes to evaluate language understanding with less constraints than SentEval, allowing users not to rely on explicit sentence embedding based models. GLUE consists of nine  classification or regression  tasks that are carried out for sentences or sentence pairs. Three tasks focus on semantic similarity, and four tasks are based on NLI, which makes GLUE arguably semantics-based, even though it also includes sentiment classification~\cite{Socher2013} and grammaticality judgment~\cite{warstadt2018neural}.

NLI can be regarded as a universal framework for evaluation. In the {\it Recast} framework~\cite{poliak2018collecting}, existing datasets (e.g. sentiment analysis) are formulated as NLI tasks. For instance, based on the sentence {\it don't waste your money}, annotated as a negative review, they use handcrafted rules to generate the following example:
({\sc premise}: {\it When asked about the product, Liam said "don't waste your money"} , {\sc hypothesis}: {\it Liam didn't like the product}, {\sc label}: entailment).
However, the generated datasets do not allow to directly measure how well a model deals with the semantic phenomena present in the original dataset, since some sentences use artificially generated reported speech. Likewise, NLI data could be used to evaluate pragmatics and discourse analysis, but it is not clear how to generate examples that are not overly artificial. Moreover, it is unclear to what extent instances in existing NLI datasets need to deal with pragmatic aspects~\cite{bowman2016thesis}.

SuperGLUE~\cite{wang-etal-2018-glue} updates GLUE with six novel tasks that are selected to be even more challenging. Two of those tasks deal with contextualized lexical semantics, another two tasks are a form of question answering, and the remaining two are NLI problems. Only one of these NLI tasks, viz. CommitmentBank~\cite{deMarneffe_Simons_Tonhauser_2019}, is related to pragmatics.

Another effort towards evaluation of general purpose NLP systems is DecaNLP~\cite{McCann2018decaNLP}. The ten tasks of this benchmark are all framed as question answering. For example, a question answering task is derived from a sentiment analysis task using artificial questions such as {\it Is this sentence positive or negative?} Four of these tasks deal with semantic parsing, and other tasks include NLI and sentiment analysis. Pragmatic phenomena can be involved in some tasks (e.g. the summarization task) although it is hard to assess to what extent.

Discourse relation prediction has punctually been used  for sentence representation learning evaluation, by~\newcite{Nie2017} and~\newcite{sileo2019discovery}, but they all used only one dataset  (viz. PDTB; Prasad et al., 2008\nocite{pdtb2.0}), which we included in our benchmark. Discourse has also been considered for evaluation in the field of machine translation. \newcite{laubli-etal-2018-machine} showed that  neural models achieve superhuman results on sentence-level translations but that current models yield underwhelming results when considering document-level translations, also making a case for discourse-aware evaluations. DiscoEval~\cite{chen-etal-2019-evaluation} proposed a more principled evaluation of discourse modeling in NLP models. However, they mirror SentEval in that they rely on sentence embeddings and fixed compositions, which has been shown to be restrictive~\cite{sileo-etal-2019-composition}, and not necessarily in line with state of the art systems. Moreover, they focus on rather shallow aspects of document structure such as the position of sentences within a document.

Other evaluations, such as linguistic probing or GLUE diagnostics~\cite{ConneauProbe,belinkov2019analysis,wang2018glue} focus on an internal understanding of what is captured by the models (e.g. syntax, lexical content), rather than measuring performance on external tasks; this provides a complementary viewpoint, but it is outside the scope of this work.


\section{PragmEval}

\subsection{Construction}

Our goal is to compile a set of diverse pragmatics-related tasks. We restrict ourselves to classification either of sentences or sentence pairs, and only use  publicly available datasets that are absent from other well-established benchmarks (such as SentEval, GLUE, and SuperGLUE), in order to have complementary benchmarks.


The scores in our tasks are not all meant to be compared to previous work, since we alter some datasets to yield more meaningful evaluations (we perform duplicate removal or class subsampling when mentioned). We found these operations necessary in order to leverage the rare classes and yield more meaningful scores. As an illustration, the  GUM discourse corpus initially consists of more than $99\% $ of {\em unattached} labels, and the dialog act annotations of the SwitchBoard conversation corpus contains $80\%$ of {\it statements}.
While disturbing the distributions of labels impacts the performance of models in real-world contexts, it seems reasonable when the goal is to indirectly evaluate the capacity of models to discriminate different semantic or pragmatic phenomena.

Section \ref{sec:pragmtasks} presents the tasks we selected, while \ref{sec:taxonomy} proposes a rudimentary taxonomy of how they address different aspects of meaning. A summary of the tasks, together with some examples, is also given in table \ref{tab:PragmEvalExamples}.

\subsection{Task overview}\label{sec:pragmtasks}



\paragraph{PDTB} The Penn Discourse Tree Bank~\cite{Prasad2014} contains a collection of fine-grained implicit (i.e. not signaled by a discourse marker) and explicit relations between sentences from the news domain in the Penn TreeBank 2.0, which signal the purpose of an utterance given a context utterance.
Explicit relations can be easily predicted from the discourse marker alone \cite{PitlerEasilyIdentifiable2008} and are discarded. 
We select the level 2 relations, called types in PDTB terminology, as categories.

\paragraph{STAC}~\cite{asher-etal-2016-discourse} is a corpus of strategic chat conversations manually annotated with negotiation-related information, dialogue acts and discourse structures in the framework of Segmented Discourse Representation Theory~\cite{asher2003logics}. We only consider pairwise relations between all dialogue acts, following~\newcite{badene-etal-2019-data}.
We remove duplicate pairs and dialogues that only have non-linguistic utterances (coming from the game server). We subsample dialogue act pairs with no relation so that they constitute $20\%$ of each fold. 

\paragraph{GUM}~\cite{Zeldes2017} is a corpus of multilayer annotations for texts from various domains; it includes discourse structure annotations according to  Rhetorical Structure Theory~\cite{mannThompson87}. Once again, we only consider pairwise interactions between discourse units (e.g. sentences/clauses). 
We subsample discourse units with no relation so that they constitute $20\%$ of each document. We split the examples in train/test/dev sets randomly according to the document they belong to.

\paragraph{Emergent}~\cite{Ferreira2016EmergentAN} is composed of pairs of assertions and titles of news articles that are {\it against}, {\it for}, or {\it neutral} with respect to the opinion of the assertion.

\paragraph{SwitchBoard}~\cite{Godfrey:1992:STS:1895550.1895693} contains textual transcriptions of dialogues about various topics with annotated speech acts. We remove duplicate examples and subsample {\it Statements} and  {\it Non Statements} so that they constitute 20\% of the examples. We use a custom train/validation split (90/10 ratio) since our preprocessing leads to a drastic size reduction of the original development set.
 The  label of a speech act can be dependent on the context (previous utterances), but we discarded it in this work for the sake of simplicity, even though integration of context could improve the scores~\cite{ribeiro2015influence}. 

\paragraph{MRDA}~\cite{shriberg2004icsi} contains textual transcriptions of multi-party real meetings, with speech act annotations. We remove duplicate examples. We use a custom train/validation split (90/10 ratio) since this deduplication leads to a drastic size reduction of the original development set, and we subsample {\it Statement} examples so that they constitute 20\% of the dataset. We also discarded the context.

\paragraph{Persuasion}~\cite{Persuasion2018Ng} is a collection of arguments from student essays  annotated with factors of persuasiveness with respect to a claim; considered factors are the following: Specificity, Eloquence, Relevance and Strength. For  each graded target, we cast the ratings into three quantiles and discard the middle quantile.

\paragraph{SarcasmV2}~\cite{OrabySarc} consists of messages from online forums with responses that may or may not be sarcastic according to human annotations.

\paragraph{Squinky dataset}~\cite{DBLP:journals/corr/Lahiri15} gathers annotations on Formality, Informativeness, and Implicature, where sentences were graded on a scale from 1 to 7.  The Implicature score is defined as the amount of information that is not explicitly expressed in a sentence. For each target, we cast the ratings into three quantiles and discard the middle quantile.

\paragraph{Verifiability}~\cite{park2014identifying} is a collection of online user comments annotated as {\it Verifiable-Experiential} (verifiable and about writer's experience),  {\it Verifiable-Non-Experiential}, or {\it Unverifiable}.

\paragraph{EmoBank}~\cite{buechel-hahn-2017-emobank} aggregates emotion annotations on texts from various domains using the VAD representation format. The authors define Valence as {\it corresponding to the concept of polarity},\footnote{This is the dimension that is widely used in sentiment analysis.} Arousal as {\it degree of calmness or excitement} and Dominance  as {\it perceived degree of control over a situation}.
For each target, we cast the ratings into three quantiles and discard the middle quantile.


\begin{table*}
\begin{center}
\begin{footnotesize}
\begin{tabular}{@{}llp{5.5cm}@{}}
\toprule
       Dataset & Example & Class \\ \midrule
          PDTB & \textit{it was censorship} / \textit{it was outrageous} & conjunction \\
          STAC & \textit{what?} / \textit{i literally lost} & question-answer-pair \\
           GUM & \textit{Do not drink} / \textit{if underage in your country} & condition \\
      Emergent & \textit{a meteorite landed in nicaragua.} / \textit{small meteorite hits managua} & for \\
   SwitchBoard & \textit{well, a little different, actually}  & hedge  \\
          MRDA & \textit{yeah that's that's that's what i meant .} & acknowledge-answer \\
    Persuasion & \textit{Co-operation is essential for team work} / \textit{lions hunt in a team} & low specificity \\
     SarcasmV2 & \textit{don't quit your day job} / \textit{[...] i was going to sell this joke. [...]} & sarcasm \\
       Squinky & \textit{boo ya.} & uninformative, high implicature, informal \\
 Verifiability & \textit{I've been a physician for 20 years.} & verifiable-experiential \\
       EmoBank & \textit{I wanted to be there..} & low valence, high arousal, low dominance \\ \bottomrule
\end{tabular}
\caption{Example instances for each of the PragmEval tasks (these examples were selected for their conciseness and are not representative of the whole dataset)
}
\label{tab:PragmEvalExamples}
\end{footnotesize}
\end{center}

\end{table*}

\subsection{Taxonomy}\label{sec:taxonomy}

It has been argued by~\newcite{Halliday85} that linguistic phenomena fall into three metafunctions: {\em ideational}
for semantics, {\it interpersonal} for appeals to the hearer/reader, and {\it textual} for form-related aspects. This forms the basis of discourse relation types by~\newcite{hovyMaier92}, who call them semantic, interpersonal and presentational. PragmEval tasks cut across these categories, because some of the tasks integrate all aspects when they characterize the speech act or discourse relation category associated to a discourse unit (mostly sentences), an utterance, or a pair of these. However, most discourse relations involved focus on {\em ideational} aspects, which are thus complemented by tasks insisting on more interpersonal aspects (e.g. using appeal to emotions, or verifiable arguments) that help realize speech act intentions. Finally, intentions can achieve their goals with varying degrees of success. This leads us to a rudimentary grouping of our tasks:

\begin{itemize}[leftmargin=*]
\item[A] The speech act classification tasks (SwitchBoard, MRDA) deal with the detection of the intention of utterances.  They use the same label set~\cite{core1997coding} but different domains and annotation guidelines. Similarly, a discourse relation characterizes how an utterance contributes to the coherence of a document/conversation (e.g through {\it elaboration} or {\it contrast}), so this task requires a form of understanding of the use of a sentence, and how a sentence fits with another sentence in a broader discourse.  A discourse relation can be seen as a speech act whose definition is tied to a structured context~\cite{asher2003logics}.  Here, three tasks (PDTB, STAC, GUM) deal with discourse relation prediction with varying  domains and formalisms.\footnote{These formalisms have different assumptions about the nature of discourse structure.} The Stance detection task can be seen as a coarse-grained discourse relation classification.

\item[B]  Detecting emotional content, verifiability, formality, informativeness or sarcasm is necessary in order to figure out in what realm communication is occurring. A statement can be persuasive, yet poorly informative and unverifiable. Emotions~\cite{dolan2002emotion} and power perception~\cite{pfeffer1981understanding}  can have a strong influence on human behavior and text interpretation. Manipulating emotions can be the main purpose of a speech act as well. Sarcasm is another means of communication and sarcasm detection is in itself a suitable task for the evaluation of pragmatics, since sarcasm is a clear case of literal meaning being different from the intended meaning.

\item[C]  Persuasiveness prediction is a useful tool to assess whether a model can measure how well a sentence can achieve its intended goal.  This aspect is orthogonal to the determination of the goal itself, and is arguably equally important.

\end{itemize}

Note that the semantic tasks of GLUE can also be considered as a grouping of tasks, where the goal is to represent accurately the denotation of utterances (e.g. the identity of the objects and agents they involve, the relation between them, the temporal and spatial location).
In contrast, solving PragmEval tasks requires knowledge of complementary aspects that characterize utterances in a different way. The A group characterizes the kind of frame into which semantic content fits; for instance, identical subjects, verbs, and objects can be used in a question, a claim, or an instruction. Semantic tasks (semantic similarity, NLI) usually compare utterances within the same frame. Additionally, utterances with the same semantic content can differ according to aspects involved in group B and C, e.g. formality or persuasiveness. To ensure that these aspects are taken into account by NLU models, a pragmatic evaluation is required.

\section{Evaluation}

\subsection{Models}
Our goal is to assess the performance of popular NLU models and the influence of various training signals on PragmEval scores.
We evaluate state of the art models and baselines on PragmEval using the Jiant framework~\cite{wang2019jiant}. Our baselines consist of an average of GloVe~\cite{Pennington} embeddings (CBoW), and a BiLSTM with both GloVe and ELMo~\cite{peters-etal-2018-deep} embeddings.
We equally evaluate BERT~\cite{devlin2018bert} base uncased  models, and perform experiments with {\it Supplementary Training on Intermediate Labeled-data Tasks}~\cite{PhangSTILTS}. STILT is a further pretraining step on a data-rich task before the final fine-tuning evaluation on the target task. STILTs can be combined using multitask learning.
We use Jiant's default parameters,\footnote{\url{https://github.com/nyu-mll/jiant/jiant/config/examples/stilts_example.conf}} and uniform loss weighting when multitasking (a different task is optimized at each training batch).

We finetune BERT with four of such training signals:
\paragraph{MNLI}~\cite{N18-1101:MNLI} is a collection of 433k sentence pairs manually annotated with {\it contradiction}, {\it entailment}, or {\it neutral} relations. \newcite{PhangSTILTS} showed that finetuning with this dataset leads to accuracy improvement on all GLUE tasks except CoLA~\cite{warstadt2018neural}.
\paragraph{DisSent}~\cite{Nie2017} consists of $4.7M$ sentence pairs that are separated by a discourse marker (from a list of 15 markers).  Prediction of discourse markers based on the context clauses/sentences with which they occur has been used as a training signal for sentence representation learning. The authors used  handcrafted rules for each marker in order to ensure that the markers signal an actual relation. DisSent has underwhelming results on the GLUE tasks as a STILT~\cite{Wang2019CanYT}.
\paragraph{Discovery}~\cite{sileo2019discovery} is another dataset for discourse marker prediction,  composed of $174$ discourse markers with $10k$ usage examples for each marker. Sentence pairs were extracted from web data, and the markers come either from the PDTB or from a heuristic automatic extraction.
\paragraph{PragmEval} refers to all PragmEval tasks used in a multitask setup; since we use a uniform loss weighting, we discard Persuasion classes other than Strength (note that the other classes can be considered subfactors for strength) in order to prevent the Persuasion task to overwhelm the others.

\subsection{Human accuracy estimates}
For a more insightful comparison, we propose derivations of human accuracy estimates for the datasets we used. The authors of SarcasmV2~\cite{OrabySarc} dataset directly report $80\%$ annotator accuracy compared to the gold standard. \newcite{Prasad2014} report $84\%$ annotator agreement for  PDTB  2.0, which is a lower bound of accuracy. For GUM~\cite{Zeldes2017}, an {\it attachment accuracy of $87.22\%$  and labelling accuracy of $86.58\%$ as compared to the `gold standard' after instructor adjudication} is reported. We interleaved attachment and labelling in our task. Assuming human annotators never predict the non-attached relation,  $69.3\%$  is a lower bound for human accuracy. Authors of the Verifiability~\cite{park2014identifying} dataset report an agreement  $\kappa=0.73$ which yields an agreement of $87\%$ given the class distribution, which is a lower bound of human accuracy. We estimated human accuracy on EmoBank~\cite{buechel-hahn-2017-emobank} with the intermediate datasets provided by the authors. For each target (V,A,D) we compute the average standard deviation, and compute the probability (under normality assumption) of each example rating of falling under the wrong category.

Unlike the GLUE benchmark~\cite{nangia-bowman-2019-human}, we do not yet provide human accuracy estimates obtained in a standardized way. The high number of classes would make that process rather more difficult. But our estimates are still useful even though they should be taken with a grain of salt.

\begin{table*}[htb]
\setlength{\tabcolsep}{2pt}
\begin{center}
\begin{footnotesize}
\noindent\makebox[\textwidth]{%

\begin{tabular}{llllllllllll}
\toprule
{} &             PDTB &             STAC &              GUM &         Emergent &         SwitchB. &             MRDA &       Persuasion &          Sarcasm &          Squinky &           Verif. &          EmoBank \\
\midrule
CBoW                &             27.4 &               32 &             20.5 &             59.7 &              3.8 &              0.7 &             70.6 &             61.1 &             75.5 &               74.0 &               64.0 \\
BiLSTM              &             25.9 &             27.7 &             18.5 &             45.6 &              3.7 &              0.7 &             62.6 &             63.1 &             72.1 &               74.0 &             63.5 \\
BiLSTM+ELMo         &             27.5 &             33.5 &             18.9 &             55.2 &              3.7 &              0.7 &             67.4 &             68.9 &             82.5 &               74.0 &             66.9 \\
\midrule
Previous work &48.2&-&-&73.1&-&-&-&-&-&81.1&-\\
\midrule
BERT                &             48.8 &             48.2 &             40.9 &             79.2 &             38.8 &             22.3 &             74.8 &  $\textbf{77.1}$ &             87.5 &             86.7 &             76.2 \\
BERT+MNLI           &             49.1 &             49.1 &             42.8 &             81.2 &             38.1 &             22.7 &             71.7 &             73.4 &             88.2 &               86.0 &             76.3 \\
BERT+PragmEval       &             49.1 &  $\textbf{57.1}$ &             42.8 &             80.2 &  $\textbf{40.3}$ &  $\textbf{23.1}$ &  $\textbf{76.2}$ &               75.0 &             87.6 &             85.9 &               76.0 \\
BERT+DisSent        &             49.4 &               49.0 &             43.9 &             79.8 &             39.2 &               22.0 &             74.7 &             74.9 &             87.5 &             85.9 &             76.2 \\
B+DisSent+MNLI   &             49.6 &             49.2 &  $\textbf{44.2}$ &             80.9 &             39.8 &             22.1 &               74.0 &             74.1 &             87.6 &             85.6 &             76.4 \\
BERT+Discovery      &             50.7 &             49.5 &             42.7 &  $\textbf{81.7}$ &             39.5 &             22.4 &             71.6 &             76.7 &             88.6 &             86.3 &  $\textbf{76.6}$ \\
B+Discovery+MNLI &  $\textbf{51.3}$ &             49.4 &             43.1 &             80.7 &  $\textbf{40.3}$ &             22.2 &             73.6 &             75.1 &  $\textbf{88.9}$ &  $\textbf{86.8}$ &               76.0 \\
\midrule
Human estimate               &  $\textbf{84.0}$ &              - &  $\textbf{69.3}$ &              - &              - &              - &              - &  $\textbf{80.0}$ &              - &  $\textbf{87.0}$ &             73.1 \\

\bottomrule
\end{tabular}
}
\caption[Transfer test accuracies across PragmEval tasks]{Transfer test scores across PragmEval tasks; we report the average when the dataset has several classification tasks (as in Squinky, EmoBank and Persuasion); 
B(ERT)+$\mathcal X$ refers to BERT pretrained classification model after an auxiliary finetuning phase on task $\mathcal X$. All scores are accuracy scores except SwitchBoard/MRDA, which are macro-F1 scores. {\it Previous work} refers to the best scores from previous work that used a similar setup, where PDTB score is from~\cite{bai-zhao-2018-deep}, Emergent score is from~\cite{Ferreira2016EmergentAN} and
Verifiability score is derived from~\cite{park2014identifying}.
}
\label{tab:transfer}
\end{footnotesize}
\end{center}
\end{table*}

\subsection{Overall results}

Task-wise results are presented in table \ref{tab:transfer}. We report the average scores of 6 runs of STILT and finetuning phases.

\noindent PragmEval seems to be challenging even for the BERT base model, which has shown strong performance on GLUE (and vastly outperforms the baselines on our tasks).  For many tasks, there is a STILT that significantly improves the accuracy of BERT.
The gap between human accuracy and BERT is particularly high on implicit discourse relation prediction (both for the PDTB corpus and the GUM RST corpus).  This task is known to be difficult, and previous work has also shown that task dedicated models are not yet on par with human performance either on PDTB~\cite{bai-zhao-2018-deep} or RST data~\cite{morey-etal-2017-much}.

Pretraining on MNLI does not improve the PragmEval average score for the BERT base model. A lower sarcasm detection score could indicate that BERT+MNLI is more focused on the literal content of statements, even though no STILT improves sarcasm detection. All models score below human accuracies, with the exception of emotion classification (but it is only due to the valence prediction subtask).

Table \ref{tab:AVG} shows aggregate results alongside comparisons with GLUE scores. The best overall unsupervised result (GLUE+PragmEval average) is achieved with Discovery STILT.
Combining Discovery and MNLI yields both a high PragmEval and GLUE score, and also yields a high GLUE diagnostics score. All discourse based STILTs improve GLUE score, while MNLI does not improve PragmEval average score. PragmEval tasks based on sentence pairs seem to account for the variance across STILTs.

MNLI has been suggested as a good default auxiliary training task based on evaluation with GLUE~\cite{PhangSTILTS} and SentEval~\cite{Conneau2017}. However, our evaluation suggests that finetuning a model with MNLI alone  has significant drawbacks.

More detailed results for datasets with several subtasks are shown in table \ref{tab:subtasks}. We note that MNLI STILT significantly decreases relevance estimation performance (on BERT base and while multi-tasking with DisSent). Many models surpass the human estimate at valence prediction, a well studied task, but interestingly this is not the case for Arousal and Dominance prediction.

\begin{table*}[h!]
\setlength{\tabcolsep}{2pt}

\begin{center}
\begin{footnotesize}
\noindent\makebox[\textwidth]{%

\begin{tabular}{llllll}
\toprule
{} &    PragmEval$_{AVG}$&     P.E.-Pairs$_{AVG}$ &   P.E.-Single$_{AVG}$&          GLUE$_{AVG}$ &   GLUE$_{diagnostics}$    \\
\midrule
BERT                &             61.8$\pm$.4 &             57.9$\pm$.5 &             62.3$\pm$.3 &             74.7$\pm$.2 &             31.7$\pm$.3 \\
BERT+MNLI           &             61.7$\pm$.5 &             57.2$\pm$.5 &             62.2$\pm$.4 &  {\bf 77.0}$\pm$.2 &             32.5$\pm$.6 \\
BERT+PragmEval MTL       &  {\bf 63.0}$\pm$.4 &  {\bf 60.0}$\pm$.4 &             62.6$\pm$.2 &             75.3$\pm$.2 &             31.6$\pm$.3 \\
BERT+DisSent        &             62.0$\pm$.4 &             58.4$\pm$.4 &             62.2$\pm$.3 &             75.1$\pm$.2 &             31.5$\pm$.3 \\
B+DisSent+MNLI   &             62.1$\pm$.4 &             58.2$\pm$.4 &             62.3$\pm$.2 &             76.6$\pm$.1 &             32.4$\pm$.0 \\
BERT+Discovery      &             62.4$\pm$.3 &             58.2$\pm$.4 &             62.7$\pm$.3 &             75.0$\pm$.2 &             31.3$\pm$.2 \\
B+Discovery+MNLI &             62.5$\pm$.4 &             58.5$\pm$.5 &             {\bf 62.8}$\pm$.3 &             76.6$\pm$.2 &  {\bf 33.3}$\pm$.2 \\
\bottomrule
\end{tabular}
}%
\caption[Aggregated transfer test accuracies across PragmEval]{ Aggregated transfer test accuracies across PragmEval  and comparison with GLUE validation downstream and diagnostic tasks (GLUE diagnostic tasks evaluate NLI performance under presence of linguistic phenomena such as negation, quantification, use of common sense);
BERT+$\mathcal X$ refers to BERT pretrained classification model after auxiliary finetuning phase on task $\mathcal X$; P.E.-Pairs$_{AVG}$ is the average of PragmEval sentence pair classification tasks.}
\label{tab:AVG}
\end{footnotesize}
\end{center}
\end{table*}

\begin{table*}[h!]
\setlength{\tabcolsep}{2pt}

\begin{center}
\begin{footnotesize}
\noindent\makebox[\textwidth]{%

\begin{tabular}{lllllllllll}
\toprule
\multicolumn{1}{c|}{}& \multicolumn{4}{c|}{Persuasiveness} & \multicolumn{3}{c|}{EmoBank}& \multicolumn{3}{c|}{Squinky} \\
{} &      Eloquence &       Relevance &     Specificity &        Strength &         Valence &         Arousal &         Dom. &  Inf. &      Implicature &        Formality \\
\midrule
BERT                &             75.6 &              63.5 &             81.6 &             78.3 &             87.1 &               72.0 &             69.5 &             92.2 &             72.1 &             98.3 \\
BERT+MNLI           &             74.7 &             57.5 &             82.3 &             72.2 &             86.6 &             72.4 &             69.9 &             92.5 &             73.9 &             98.1 \\
BERT+PragmEval       &             75.6 &               {\bf 64.0} &             83.2 &  $\textbf{82.0}$ &             86.8 &             71.9 &             69.2 &             92.3 &             71.8 &  $\textbf{98.6}$ \\
BERT+DisSent        &             73.8 &               63.0 &             82.6 &             79.5 &             87.1 &             71.4 &             70.1 &             92.6 &               72.0 &             97.7 \\
B+DisSent+MNLI   &  $\textbf{76.9}$ &             61.5 &  $\textbf{83.9}$ &             73.9 &  $\textbf{87.6}$ &             72.1 &             69.4 &             91.5 &             73.4 &             97.9 \\
BERT+Discovery      &               76.0 &             59.1 &             80.1 &             71.4 &             86.8 &  $\textbf{72.6}$ &  $\textbf{70.5}$ &  $\textbf{93.2}$ &             74.2 &             98.5 \\
B+Discovery+MNLI &             74.1 &             60.4 &             79.4 &             80.4 &             86.4 &             72.1 &             69.6 &             93.1 &  $\textbf{75.3}$ &             98.4 \\
\midrule
Human estimate               &              - &              - &              - &              - &             74.9 &  $\textbf{73.8}$ &            {\bf 70.5} &              - &              - &              - \\
\bottomrule
\end{tabular}
}%
\caption[Transfer test accuracies across a few PragmEval subtasks]{Transfer test accuracies across PragmEval subtasks (Persuasiveness, EmoBank, Squinky)
BERT+$\mathcal X$ refers to BERT pretrained classification model after auxiliary finetuning phase on task $\mathcal X$.}
\label{tab:subtasks}
\end{footnotesize}
\end{center}
\end{table*}


The categories of our benchmark tasks cover a broad range of pragmatic aspects. The overall accuracies only show a synthetic view of the tasks evaluated in PragmEval. Some datasets contain many subcategories that allow for a fine grained analysis through a wide array of classes (e.g. 51 categories for MRDA).
Table \ref{tab:finegrained} in appendix A shows a fine grained evaluation which yields some insights on the capabilities of BERT. We report the 5 most frequent classes per task. It is worth noting that the BERT models do not neglect rare classes. These detailed results reveal that BERT+MNLI scores for discourse relation prediction are inflated by good scores on predicting the absence of relation (possibly close to the neutral class in NLI), which is useful but not sufficient for pragmatics understanding.
The STILTs have complementary strengths even with given tasks, which can explain why combining them is helpful. However, we used a rather simplistic multitask setup, and efficient combination of the tasks remains an open problem.

\section{Conclusion}
We proposed PragmEval, a set of pragmatics related evaluation tasks, and used them to evaluate BERT finetuned on various auxiliary finetuning tasks.  The results lead us to rethink the efficiency of mainly using NLI as an auxiliary training task.
PragmEval can be used for training or evaluating NLU or pragmatics related work in general. Much effort has been devoted to NLI for training and evaluation for general purpose sentence understanding, but we just scratched the surface of the use of pragmatics oriented tasks.
In further investigations, we plan to use more general tasks than classification on sentences or sentence pairs, such as longer and possibly structured sequences.  Several of the datasets we used (MRDA, SwitchBoard, GUM, STAC) already contain such higher level structures.
Of course defining a generic architecture for structured tasks in which to evaluate the contribution of trained representations is not straightforward. 
In addition, a more inclusive comparison with human annotators on pragmatics tasks could also help to pinpoint the weaknesses of current models dealing with pragmatics phenomena. Yet another step would be to study the correlations between performance metrics in deployed NLU systems and scores of the automated evaluation benchmarks (GLUE/PragmEval) in order to validate our claims about the centrality of pragmatics.


\nocite{sileo-etal-2020-discsense}

\bibliographystyle{lrec2022-bib}
\bibliography{main}

\begin{thebibliography}{}

\bibitem[\protect\citename{Agirre \bgroup et al.\egroup
  }2012]{agirre2012semeval}
Agirre, E., Diab, M., Cer, D., and Gonzalez-Agirre, A.
\newblock (2012).
\newblock Semeval-2012 task 6: A pilot on semantic textual similarity.
\newblock In {\em Proceedings of the First Joint Conference on Lexical and
  Computational Semantics -- Volume 2: Proceedings of the Sixth International
  Workshop on Semantic Evaluation}, pages 385--393. Association for
  Computational Linguistics.

\bibitem[\protect\citename{Asher and Lascarides}2003]{asher2003logics}
Asher, N. and Lascarides, A.
\newblock (2003).
\newblock {\em Logics of conversation}.
\newblock Cambridge University Press.

\bibitem[\protect\citename{Asher \bgroup et al.\egroup
  }2016]{asher-etal-2016-discourse}
Asher, N., Hunter, J., Morey, M., Farah, B., and Afantenos, S.
\newblock (2016).
\newblock Discourse structure and dialogue acts in multiparty dialogue: the
  {STAC} corpus.
\newblock In {\em Proceedings of the Tenth International Conference on Language
  Resources and Evaluation ({LREC}'16)}, pages 2721--2727, Portoro{\v{z}},
  Slovenia, May. European Language Resources Association (ELRA).

\bibitem[\protect\citename{Austin}1975]{austin1975things}
Austin, J.~L.
\newblock (1975).
\newblock {\em How to do things with words}.
\newblock Oxford university press.

\bibitem[\protect\citename{Badene \bgroup et al.\egroup
  }2019]{badene-etal-2019-data}
Badene, S., Thompson, K., Lorr{\'e}, J.-P., and Asher, N.
\newblock (2019).
\newblock Data programming for learning discourse structure.
\newblock In {\em Proceedings of the 57th Conference of the Association for
  Computational Linguistics}, pages 640--645, Florence, Italy, July.
  Association for Computational Linguistics.

\bibitem[\protect\citename{Bai and Zhao}2018]{bai-zhao-2018-deep}
Bai, H. and Zhao, H.
\newblock (2018).
\newblock Deep enhanced representation for implicit discourse relation
  recognition.
\newblock In {\em Proceedings of the 27th International Conference on
  Computational Linguistics}, pages 571--583, Santa Fe, New Mexico, USA,
  August. Association for Computational Linguistics.

\bibitem[\protect\citename{Belinkov and Glass}2019]{belinkov2019analysis}
Belinkov, Y. and Glass, J.
\newblock (2019).
\newblock Analysis methods in neural language processing: A survey.
\newblock {\em Transactions of the Association for Computational Linguistics},
  7:49--72.

\bibitem[\protect\citename{Bowman}2016]{bowman2016thesis}
Bowman, S.~R.
\newblock (2016).
\newblock {\em {Modeling natural language semantics in learned
  representations}}.
\newblock {Ph.D.} thesis.

\bibitem[\protect\citename{Buechel and Hahn}2017]{buechel-hahn-2017-emobank}
Buechel, S. and Hahn, U.
\newblock (2017).
\newblock {E}mo{B}ank: Studying the impact of annotation perspective and
  representation format on dimensional emotion analysis.
\newblock In {\em Proceedings of the 15th Conference of the {E}uropean Chapter
  of the Association for Computational Linguistics: Volume 2, Short Papers},
  pages 578--585, Valencia, Spain, April. Association for Computational
  Linguistics.

\bibitem[\protect\citename{Carlile \bgroup et al.\egroup
  }2018]{Persuasion2018Ng}
Carlile, W., Gurrapadi, N., Ke, Z., and Ng, V.
\newblock (2018).
\newblock Give me more feedback: Annotating argument persuasiveness and related
  attributes in student essays.
\newblock In {\em Proceedings of the 56th Annual Meeting of the Association for
  Computational Linguistics (Volume 1: Long Papers)}, pages 621--631,
  Melbourne, Australia, July. Association for Computational Linguistics.

\bibitem[\protect\citename{Cer \bgroup et al.\egroup }2018]{cer2018universal}
Cer, D., Yang, Y., yi~Kong, S., Hua, N., Limtiaco, N., John, R.~S., Constant,
  N., Guajardo-Cespedes, M., Yuan, S., Tar, C., Sung, Y.-H., Strope, B., and
  Kurzweil, R.
\newblock (2018).
\newblock Universal sentence encoder.

\bibitem[\protect\citename{Chen \bgroup et al.\egroup
  }2019]{chen-etal-2019-evaluation}
Chen, M., Chu, Z., and Gimpel, K.
\newblock (2019).
\newblock Evaluation benchmarks and learning criteria for discourse-aware
  sentence representations.
\newblock In {\em Proceedings of the 2019 Conference on Empirical Methods in
  Natural Language Processing and the 9th International Joint Conference on
  Natural Language Processing (EMNLP-IJCNLP)}, pages 649--662, Hong Kong,
  China, November. Association for Computational Linguistics.

\bibitem[\protect\citename{Conneau \bgroup et al.\egroup }2017]{Conneau2017}
Conneau, A., Kiela, D., Schwenk, H., Barrault, L., and Bordes, A.
\newblock (2017).
\newblock {Supervised Learning of Universal Sentence Representations from
  Natural Language Inference Data}.
\newblock {\em Emnlp}.

\bibitem[\protect\citename{Conneau \bgroup et al.\egroup }2018]{ConneauProbe}
Conneau, A., Kruszewski, G., Lample, G., Barrault, L., and Baroni, M.
\newblock (2018).
\newblock What you can cram into a single vector: Probing sentence embeddings
  for linguistic properties.
\newblock In {\em Proceedings of the 56th Annual Meeting of the Association for
  Computational Linguistics (Volume 1: Long Papers)}, pages 2126--2136.
  Association for Computational Linguistics.

\bibitem[\protect\citename{Core and Allen}1997]{core1997coding}
Core, M.~G. and Allen, J.
\newblock (1997).
\newblock Coding dialogs with the {DAMSL} annotation scheme.
\newblock In {\em AAAI fall symposium on communicative action in humans and
  machines}, volume~56, pages 28--35. Boston, MA.

\bibitem[\protect\citename{de Marneffe \bgroup et al.\egroup
  }2019]{deMarneffe_Simons_Tonhauser_2019}
de~Marneffe, M.-C., Simons, M., and Tonhauser, J.
\newblock (2019).
\newblock The commitmentbank: Investigating projection in naturally occurring
  discourse.
\newblock {\em Proceedings of Sinn und Bedeutung}, 23(2):107--124, Jul.

\bibitem[\protect\citename{Devlin \bgroup et al.\egroup }2019]{devlin2018bert}
Devlin, J., Chang, M.-W., Lee, K., and Toutanova, K.
\newblock (2019).
\newblock Bert: Pre-training of deep bidirectional transformers for language
  understanding.
\newblock In {\em Proceedings of the 2019 Conference of the North American
  Chapter of the Association for Computational Linguistics: Human Language
  Technologies, Volume 1 (Long Papers)}. Association for Computational
  Linguistics.

\bibitem[\protect\citename{Dolan}2002]{dolan2002emotion}
Dolan, R.~J.
\newblock (2002).
\newblock Emotion, cognition, and behavior.
\newblock {\em Science}, 298(5596):1191--1194.

\bibitem[\protect\citename{Ferreira and Vlachos}2016]{Ferreira2016EmergentAN}
Ferreira, W. and Vlachos, A.
\newblock (2016).
\newblock Emergent: a novel data-set for stance classification.
\newblock In {\em HLT-NAACL}.

\bibitem[\protect\citename{Godfrey \bgroup et al.\egroup
  }1992]{Godfrey:1992:STS:1895550.1895693}
Godfrey, J.~J., Holliman, E.~C., and McDaniel, J.
\newblock (1992).
\newblock Switchboard: Telephone speech corpus for research and development.
\newblock In {\em Proceedings of the 1992 IEEE International Conference on
  Acoustics, Speech and Signal Processing - Volume 1}, ICASSP'92, pages
  517--520, Washington, DC, USA. IEEE Computer Society.

\bibitem[\protect\citename{Green}2000]{GreenForceAndContent}
Green, M.~S.
\newblock (2000).
\newblock Illocutionary force and semantic content.
\newblock {\em Linguistics and Philosophy}, 23(5):435--473.

\bibitem[\protect\citename{Grice}1975]{grice1975logic}
Grice, H.~P.
\newblock (1975).
\newblock Logic and conversation.
\newblock In Peter Cole et~al., editors, {\em Syntax and Semantics: Vol. 3:
  Speech Acts}, pages 41--58. Academic Press, New York.

\bibitem[\protect\citename{Halliday}1985]{Halliday85}
Halliday, M.
\newblock (1985).
\newblock {\em An Introduction to Functional Grammar}.
\newblock Edward Arnold Press, Baltimore.

\bibitem[\protect\citename{Hovy and Maier}1992]{hovyMaier92}
Hovy, E. and Maier, E.
\newblock (1992).
\newblock Parsimonious or profligate: How many and which discourse structure
  relations?
\newblock Technical Report RR-93-373, USC Information Sciences Institute.

\bibitem[\protect\citename{Kiros and Chan}2018]{kiros-chan-2018-inferlite}
Kiros, J. and Chan, W.
\newblock (2018).
\newblock {I}nfer{L}ite: Simple universal sentence representations from natural
  language inference data.
\newblock In {\em Proceedings of the 2018 Conference on Empirical Methods in
  Natural Language Processing}, pages 4868--4874, Brussels, Belgium,
  October-November. Association for Computational Linguistics.

\bibitem[\protect\citename{Kiros \bgroup et al.\egroup }2015]{Kiros2015}
Kiros, R., Zhu, Y., Salakhutdinov, R.~R., Zemel, R., Urtasun, R., Torralba, A.,
  and Fidler, S.
\newblock (2015).
\newblock {Skip-thought vectors}.
\newblock In {\em Advances in neural information processing systems}, pages
  3294--3302.

\bibitem[\protect\citename{Lahiri}2015]{DBLP:journals/corr/Lahiri15}
Lahiri, S.
\newblock (2015).
\newblock {SQUINKY! A Corpus of Sentence-level Formality, Informativeness, and
  Implicature}.
\newblock {\em CoRR}, abs/1506.02306.

\bibitem[\protect\citename{L{\"a}ubli \bgroup et al.\egroup
  }2018]{laubli-etal-2018-machine}
L{\"a}ubli, S., Sennrich, R., and Volk, M.
\newblock (2018).
\newblock Has machine translation achieved human parity? a case for
  document-level evaluation.
\newblock In {\em Proceedings of the 2018 Conference on Empirical Methods in
  Natural Language Processing}, pages 4791--4796, Brussels, Belgium,
  October-November. Association for Computational Linguistics.

\bibitem[\protect\citename{Liu \bgroup et al.\egroup }2019]{liu2019mt-dnn-kd}
Liu, X., He, P., Chen, W., and Gao, J.
\newblock (2019).
\newblock Improving multi-task deep neural networks via knowledge distillation
  for natural language understanding.
\newblock {\em arXiv preprint arXiv:1904.09482}.

\bibitem[\protect\citename{Mann and Thompson}1987]{mannThompson87}
Mann, W. and Thompson, S.
\newblock (1987).
\newblock Rhetorical structure theory : a theory of text organization.
\newblock Technical report, Information Science Institute.

\bibitem[\protect\citename{McCann \bgroup et al.\egroup
  }2018]{McCann2018decaNLP}
McCann, B., Keskar, N.~S., Xiong, C., and Socher, R.
\newblock (2018).
\newblock The natural language decathlon: Multitask learning as question
  answering.
\newblock {\em arXiv preprint arXiv:1806.08730}.

\bibitem[\protect\citename{Morey \bgroup et al.\egroup
  }2017]{morey-etal-2017-much}
Morey, M., Muller, P., and Asher, N.
\newblock (2017).
\newblock How much progress have we made on {RST} discourse parsing? a
  replication study of recent results on the {RST}-{DT}.
\newblock In {\em Proceedings of the 2017 Conference on Empirical Methods in
  Natural Language Processing}, pages 1319--1324, Copenhagen, Denmark,
  September. Association for Computational Linguistics.

\bibitem[\protect\citename{Nangia and Bowman}2019]{nangia-bowman-2019-human}
Nangia, N. and Bowman, S.~R.
\newblock (2019).
\newblock Human vs. muppet: A conservative estimate of human performance on the
  {GLUE} benchmark.
\newblock In {\em Proceedings of the 57th Annual Meeting of the Association for
  Computational Linguistics}, pages 4566--4575, Florence, Italy, July.
  Association for Computational Linguistics.

\bibitem[\protect\citename{Nie \bgroup et al.\egroup }2019]{Nie2017}
Nie, A., Bennett, E.~D., and Goodman, N.~D.
\newblock (2019).
\newblock {DisSent: Sentence Representation Learning from Explicit Discourse
  Relations}.
\newblock pages 4497--4510, July.

\bibitem[\protect\citename{Oraby \bgroup et al.\egroup }2016]{OrabySarc}
Oraby, S., Harrison, V., Reed, L., Hernandez, E., Riloff, E., and Walker, M.
\newblock (2016).
\newblock Creating and characterizing a diverse corpus of sarcasm in dialogue.
\newblock In {\em Proceedings of the 17th Annual Meeting of the Special
  Interest Group on Discourse and Dialogue}, pages 31--41. Association for
  Computational Linguistics.

\bibitem[\protect\citename{Park and Cardie}2014]{park2014identifying}
Park, J. and Cardie, C.
\newblock (2014).
\newblock Identifying appropriate support for propositions in online user
  comments.
\newblock In {\em Proceedings of the first workshop on argumentation mining},
  pages 29--38.

\bibitem[\protect\citename{Pennington \bgroup et al.\egroup }2014]{Pennington}
Pennington, J., Socher, R., and Manning, C.~D.
\newblock (2014).
\newblock {GloVe: Global Vectors for Word Representation}.
\newblock {\em Proceedings of the 2014 Conference on Empirical Methods in
  Natural Language Processing}, pages 1532--1543.

\bibitem[\protect\citename{Peters \bgroup et al.\egroup
  }2018]{peters-etal-2018-deep}
Peters, M., Neumann, M., Iyyer, M., Gardner, M., Clark, C., Lee, K., and
  Zettlemoyer, L.
\newblock (2018).
\newblock Deep contextualized word representations.
\newblock In {\em Proceedings of the 2018 Conference of the North {A}merican
  Chapter of the Association for Computational Linguistics: Human Language
  Technologies, Volume 1 (Long Papers)}, pages 2227--2237, New Orleans,
  Louisiana, June. Association for Computational Linguistics.

\bibitem[\protect\citename{Pfeffer}1981]{pfeffer1981understanding}
Pfeffer, J.
\newblock (1981).
\newblock Understanding the role of power in decision making.
\newblock {\em Jay M. Shafritz y J. Steven Ott, Classics of Organization
  Theory, Wadsworth}, pages 137--154.

\bibitem[\protect\citename{Phang \bgroup et al.\egroup }2018]{PhangSTILTS}
Phang, J., F{\'{e}}vry, T., and Bowman, S.~R.
\newblock (2018).
\newblock Sentence encoders on stilts: Supplementary training on intermediate
  labeled-data tasks.
\newblock {\em CoRR}, abs/1811.01088.

\bibitem[\protect\citename{Pitler \bgroup et al.\egroup
  }2008]{PitlerEasilyIdentifiable2008}
Pitler, E., Raghupathy, M., Mehta, H., Nenkova, A., Lee, A., and Joshi, A.
\newblock (2008).
\newblock Easily identifiable discourse relations.
\newblock In {\em Coling 2008: Companion volume: Posters}, pages 87--90. Coling
  2008 Organizing Committee.

\bibitem[\protect\citename{Poliak \bgroup et al.\egroup
  }2018]{poliak2018collecting}
Poliak, A., Haldar, A., Rudinger, R., Hu, J.~E., Pavlick, E., White, A.~S., and
  Van~Durme, B.
\newblock (2018).
\newblock Collecting diverse natural language inference problems for sentence
  representation evaluation.
\newblock In {\em Proceedings of the 2018 Conference on Empirical Methods in
  Natural Language Processing}, pages 67--81.

\bibitem[\protect\citename{Prasad \bgroup et al.\egroup }2008]{pdtb2.0}
Prasad, R., Dinesh, N., Lee, A., Miltsakaki, E., Robaldo, L., Joshi, A., and
  Webber, B.
\newblock (2008).
\newblock The penn discourse treebank 2.0.
\newblock In Nicoletta Calzolari, et~al., editors, {\em Proceedings of the
  Sixth International Conference on Language Resources and Evaluation
  (LREC'08)}, Marrakech, Morocco, may. European Language Resources Association
  (ELRA).
\newblock http://www.lrec-conf.org/proceedings/lrec2008/.

\bibitem[\protect\citename{Prasad \bgroup et al.\egroup }2014]{Prasad2014}
Prasad, R., Riley, K.~F., and Lee, A.
\newblock (2014).
\newblock {Towards Full Text Shallow Discourse Relation Annotation :
  Experiments with Cross-Paragraph Implicit Relations in the PDTB}.
\newblock (2009).

\bibitem[\protect\citename{Ribeiro \bgroup et al.\egroup
  }2015]{ribeiro2015influence}
Ribeiro, E., Ribeiro, R., and de~Matos, D.~M.
\newblock (2015).
\newblock The influence of context on dialogue act recognition.
\newblock {\em arXiv preprint arXiv:1506.00839}.

\bibitem[\protect\citename{Searle \bgroup et al.\egroup
  }1980]{searle1980speech}
Searle, J.~R., Kiefer, F., Bierwisch, M., et~al.
\newblock (1980).
\newblock {\em Speech act theory and pragmatics}, volume~10.
\newblock Springer.

\bibitem[\protect\citename{Shriberg \bgroup et al.\egroup
  }2004]{shriberg2004icsi}
Shriberg, E., Dhillon, R., Bhagat, S., Ang, J., and Carvey, H.
\newblock (2004).
\newblock The icsi meeting recorder dialog act (mrda) corpus.
\newblock In {\em Proceedings of the 5th SIGdial Workshop on Discourse and
  Dialogue at HLT-NAACL 2004}.

\bibitem[\protect\citename{Sileo \bgroup et al.\egroup
  }2019a]{sileo-etal-2019-composition}
Sileo, D., Van De~Cruys, T., Pradel, C., and Muller, P.
\newblock (2019a).
\newblock Composition of sentence embeddings: Lessons from statistical
  relational learning.
\newblock In {\em Proceedings of the Eighth Joint Conference on Lexical and
  Computational Semantics (*{SEM} 2019)}, pages 33--43, Minneapolis, Minnesota,
  June. Association for Computational Linguistics.

\bibitem[\protect\citename{Sileo \bgroup et al.\egroup
  }2019b]{sileo2019discovery}
Sileo, D., {Van de Cruys}, T., Pradel, C., and Muller, P.
\newblock (2019b).
\newblock Mining discourse markers for unsupervised sentence representation
  learning.
\newblock In {\em Proceedings of the 2019 Conference of the North American
  Chapter of the Association for Computational Linguistics: Human Language
  Technologies, Volume 1 (Long Papers)}. Association for Computational
  Linguistics.

\bibitem[\protect\citename{Sileo \bgroup et al.\egroup
  }2020]{sileo-etal-2020-discsense}
Sileo, D., Van~de Cruys, T., Pradel, C., and Muller, P.
\newblock (2020).
\newblock {D}isc{S}ense: Automated semantic analysis of discourse markers.
\newblock In {\em Proceedings of the 12th Language Resources and Evaluation
  Conference}, pages 991--999, Marseille, France, May. European Language
  Resources Association.

\bibitem[\protect\citename{Socher \bgroup et al.\egroup }2013]{Socher2013}
Socher, R., Chen, D., Manning, C., Chen, D., and Ng, A.
\newblock (2013).
\newblock {Reasoning With Neural Tensor Networks for Knowledge Base
  Completion}.
\newblock In {\em Neural Information Processing Systems (2003)}, pages
  926--934.

\bibitem[\protect\citename{Subramanian \bgroup et al.\egroup
  }2018]{subramanian2018learning}
Subramanian, S., Trischler, A., Bengio, Y., and Pal, C.~J.
\newblock (2018).
\newblock Learning general purpose distributed sentence representations via
  large scale multi-task learning.
\newblock {\em International Conference on Learning Representations}.

\bibitem[\protect\citename{Wang \bgroup et al.\egroup
  }2018]{wang-etal-2018-glue}
Wang, A., Singh, A., Michael, J., Hill, F., Levy, O., and Bowman, S.
\newblock (2018).
\newblock {GLUE}: A multi-task benchmark and analysis platform for natural
  language understanding.
\newblock In {\em Proceedings of the 2018 {EMNLP} Workshop {B}lackbox{NLP}:
  Analyzing and Interpreting Neural Networks for {NLP}}, pages 353--355,
  Brussels, Belgium, November. Association for Computational Linguistics.

\bibitem[\protect\citename{Wang \bgroup et al.\egroup }2019a]{Wang2019CanYT}
Wang, A., Hula, J., Xia, P., Pappagari, R., McCoy, R.~T., Patel, R., Kim, N.,
  Tenney, I., Huang, Y., Yu, K., Jin, S., Chen, B., Durme, B.~V., Grave, E.,
  Pavlick, E., and Bowman, S.~R.
\newblock (2019a).
\newblock Can you tell me how to get past sesame street? sentence-level
  pretraining beyond language modeling.
\newblock In {\em ACL 2019}.

\bibitem[\protect\citename{Wang \bgroup et al.\egroup }2019b]{wang2018glue}
Wang, A., Singh, A., Michael, J., Hill, F., Levy, O., and Bowman, S.~R.
\newblock (2019b).
\newblock {GLUE}: A multi-task benchmark and analysis platform for natural
  language understanding.
\newblock In {\em International Conference on Learning Representations}.

\bibitem[\protect\citename{Wang \bgroup et al.\egroup }2019c]{wang2019jiant}
Wang, A., Tenney, I.~F., Pruksachatkun, Y., Yu, K., Hula, J., Xia, P.,
  Pappagari, R., Jin, S., McCoy, R.~T., Patel, R., Huang, Y., Phang, J., Grave,
  E., Liu, H., Kim, N., Htut, P.~M., F'{e}vry, T., Chen, B., Nangia, N.,
  Mohananey, A., Kann, K., Bordia, S., Patry, N., Benton, D., Pavlick, E., and
  Bowman, S.~R.
\newblock (2019c).
\newblock \texttt{jiant} 1.2: A software toolkit for research on
  general-purpose text understanding models.
\newblock \url{http://jiant.info/}.

\bibitem[\protect\citename{Warstadt \bgroup et al.\egroup
  }2018]{warstadt2018neural}
Warstadt, A., Singh, A., and Bowman, S.~R.
\newblock (2018).
\newblock Neural network acceptability judgments.
\newblock {\em arXiv preprint arXiv:1805.12471}.

\bibitem[\protect\citename{Wieting \bgroup et al.\egroup
  }2015]{wieting2016iclr}
Wieting, J., Bansal, M., Gimpel, K., and Livescu, K.
\newblock (2015).
\newblock Towards universal paraphrastic sentence embeddings.
\newblock {\em CoRR}, abs/1511.08198.

\bibitem[\protect\citename{Williams \bgroup et al.\egroup }2018]{N18-1101:MNLI}
Williams, A., Nangia, N., and Bowman, S.
\newblock (2018).
\newblock A broad-coverage challenge corpus for sentence understanding through
  inference.
\newblock In {\em Proceedings of the 2018 Conference of the North American
  Chapter of the Association for Computational Linguistics: Human Language
  Technologies, Volume 1 (Long Papers)}, pages 1112--1122. Association for
  Computational Linguistics.

\bibitem[\protect\citename{Yang \bgroup et al.\egroup }2019]{yang2019xlnet}
Yang, Z., Dai, Z., Yang, Y., Carbonell, J., Salakhutdinov, R., and Le, Q.~V.
\newblock (2019).
\newblock Xlnet: Generalized autoregressive pretraining for language
  understanding.

\bibitem[\protect\citename{Zeldes}2017]{Zeldes2017}
Zeldes, A.
\newblock (2017).
\newblock The {GUM} corpus: Creating multilayer resources in the classroom.
\newblock {\em Language Resources and Evaluation}, 51(3):581--612.

\end{thebibliography}

\clearpage

\appendix

\onecolumn
\section*{Appendix A}

\begin{table}[h]
\setlength{\tabcolsep}{2pt}
\begin{center}
\begin{footnotesize}
\noindent\makebox[\textwidth]{%

\begin{tabular}{llllllr}
\toprule
{} &        BERT &      B+MNLI &   B+DisSent & B+Discovery &  B+PragmEval &  Support \\
\midrule
GUM.no\_relation                   &        48.9 &  {\bf 51.0} &          46.0 &        45.4 &        43.3 &       48 \\
GUM.circumstance                  &        77.1 &  {\bf 80.6} &        73.2 &        77.8 &        74.6 &       35 \\
GUM.elaboration                   &        41.5 &        38.5 &          40.0 &  {\bf 46.1} &        42.9 &       32 \\
STAC.no\_relation                  &        59.9 &  {\bf 63.8} &        55.4 &        61.3 &        46.9 &      117 \\
STAC.Comment                      &        77.8 &        76.1 &        74.9 &  {\bf 78.6} &        54.4 &      115 \\
STAC.Question\_answer\_pair         &        79.1 &        80.1 &  {\bf 83.3} &        76.9 &          83.0 &       93 \\
SwitchBoard.Uninterpretable       &          86.0 &          86.0 &        85.5 &        86.1 &  {\bf 86.3} &      382 \\
SwitchBoard.Statement-non-opinion &          72.0 &        72.1 &  {\bf 72.4} &  {\bf 72.4} &  {\bf 72.4} &      304 \\
SwitchBoard.Yes-No-Question       &  {\bf 85.9} &        85.2 &        85.5 &  {\bf 85.9} &        85.8 &      303 \\
PDTB.Cause                       &        55.2 &        55.7 &        53.1 &  {\bf 57.2} &        55.9 &      302 \\
PDTB.Restatement                 &        40.4 &          40.0 &        41.3 &  {\bf 43.9} &          41.0 &      263 \\
PDTB.Conjunction                 &        52.8 &  {\bf 53.9} &        52.1 &        53.3 &        52.5 &      262 \\
MRDA.Statement                    &        51.2 &        51.8 &        48.9 &  {\bf 53.4} &        51.4 &      364 \\
MRDA.Defending/Explanation        &        52.8 &        54.1 &  {\bf 55.3} &        52.8 &          52.0 &      166 \\
MRDA.Expansions of y/n Answers    &  {\bf 51.7} &        48.7 &        50.3 &        49.6 &        49.4 &      139 \\
\bottomrule
\end{tabular}

}%

\caption[Transfer F1 scores across some categories of PragmEval tasks]{
Transfer F1 scores across the categories of PragmEval tasks;
B(ERT)+$\mathcal{X}$ denotes BERT pretrained classification model after auxiliary finetuning phase on task $\mathcal X$.
}
\label{tab:finegrained}
\end{footnotesize}
\end{center}
\end{table}


\end{document}